\documentclass{article}
\usepackage{spconf,amsmath,graphicx,amsfonts}
\usepackage{algorithmic}
\usepackage{multirow}
\usepackage{balance}
\usepackage{bbding}
\usepackage{threeparttable}

\title {mmBaT: A multi-task framework for mmwave-based human body reconstruction and translation prediction}
\name{Jiarui Yang, Songpengcheng Xia, Yifan Song, Qi Wu, Ling Pei*
\thanks{The work was supported by the National Natural Science Foundation of China under Grant 62273229. *Corresponding author: Ling Pei: \texttt{ling.pei@sjtu.edu.cn}}} 
\address{
School of Electronic Information and Electrical Engineering,\\ Shanghai Jiao Tong University, Shanghai, China}

\begin{document}
\ninept
\maketitle
\begin{abstract}
Human body reconstruction with Millimeter Wave (mmWave) radar point clouds has gained significant interest due to its ability to work in adverse environments and its capacity to mitigate privacy concerns associated with traditional camera-based solutions. Despite pioneering efforts in this field, two challenges persist. Firstly, raw point clouds contain massive noise points, usually caused by the ambient objects and multi-path effects of Radio Frequency (RF) signals. Recent approaches typically rely on prior knowledge or elaborate preprocessing methods, limiting their applicability. Secondly, even after noise removal, the sparse and inconsistent body-related points pose an obstacle to accurate human body reconstruction. To address these challenges, we introduce mmBaT, a novel multi-task deep learning framework that concurrently estimates the human body and predicts body translations in subsequent frames to extract body-related point clouds.
Our method is evaluated on two public datasets that are collected with different radar devices and noise levels. A comprehensive comparison against other state-of-the-art methods demonstrates our method has a superior reconstruction performance and generalization ability from noisy raw data, even when compared to methods provided with body-related point clouds.
\end{abstract}

\section{Introduction}
\label{sec:intro}
Wireless sensing technology has been gaining increasingly popularity in various human-centric research domains, including vital signs monitoring \cite{1,2}, human action recognition \cite{xia2022boundary, 3} and pose estimation \cite{5,7}. Despite the extensive progress in traditional camera-based \cite{tian2023recovering} and IMU-based \cite{pei2021mars} solutions, wireless sensors have great advantages for the non-invasive and privacy-preserving attributes. 

Owing to its superior range and angle resolution, mmWave radar is gradually becoming a preferred choice for human-centric tasks, where human body reconstruction from mmWave radar point clouds has gained significant interest from both academia and industry. Sengupta et al. \cite{sengupta2020mm} utilize two commercial mmWave radars to get point clouds, using a forked 3D Convolutional Neural Network (CNN) architecture to estimate 25 skeletal 3D joints. Xue et al. introduce mmMesh \cite{xue2021mmesh}, the first approach to regress the Skinned Multi-Person Linear (SMPL) model \cite{bogo2016SMPL} body parameters while classifying gender simultaneously. In \cite{xue2022m4esh}, the authors successfully perform body reconstruction for three different subjects in a top-down fashion, separating each subject's point clouds for individual processing. Chen et al. \cite{chen2023immfusion} pioneer the fusion of mmWave radar and RGB camera data, enhancing reconstruction results in challenging environments such as darkness, rain, and fog. \par
To extract body-related information from raw mmWave radar point clouds, elaborate preprocessing methods or prior knowledge are often a prerequisite to filter out noise points.
Consequently, adopting these preprocessing methods across different radar devices and varying environmental conditions presents substantial challenges. For example, in mmMesh approach \cite{xue2021mmesh}, the range between the human body and the radar is assumed to be known, and researchers select points with the highest energy within this specified range. In \cite{xue2022m4esh}, researchers initially locate each subject during the Intermediate Frequency (IF) signal processing stage, after which they generate the body-related point cloud individually. In \cite{chen2023immfusion}, the ground-truth bounding box is used to filter out the noise points. Thus, we have to design a robust strategy to extract body-related points without any prior knowledge or preprocessing methods. 
\par
Furthermore, another significant challenge lies in the regression of the SMPL pose and shape parameters from the sparse and inconsistent point clouds, which typically exhibit an implicit and highly nonlinear character \cite{22}. This challenge poses a fundamental question in various fields that has raised significant attention in research over recent years. Cho et al. \cite{cho2021rethinking} design a hierarchical network architecture for image deblurring, which first deblurs the input image in multiple scales, then the features of the multi-scale images are fused to output the final result. To achieve accurate localization on a large scale, Sarlin et al. \cite{sarlin2019coarse} adopt a two-step method, a few coarse location hypotheses are proposed in the first step, and the local features are utilized to map with these candidate locations to find the real one. Inspired by the coarse-to-fine techniques employed in these previous studies, we adopt a novel approach by incorporating a coarse skeleton as an intermediate step in our pipeline instead of directly regressing the SMPL parameters. This selection is motivated by the robust spatial correlation that exists between the coarse skeleton and the point cloud data, which can provide a rich prior for the reconstruction of the human body, especially for the pose parameters.
\par
To address the aforementioned challenges, we propose a multi-task framework that concurrently estimates human body pose and shape while predicting subsequent frame translations for the subject. The predicted translations provide the subject's bounding boxes, facilitating the selection of body-related points for the next time window. The reconstruction of the human body is executed in a coarse-to-fine approach: an initial coarse skeleton is estimated from the point cloud data, after which we merge the features of the skeleton joints with point cloud features to regress the final body parameters.
\par
\begin{figure*}[htb]
\centering
\centerline{\includegraphics[width=17cm]{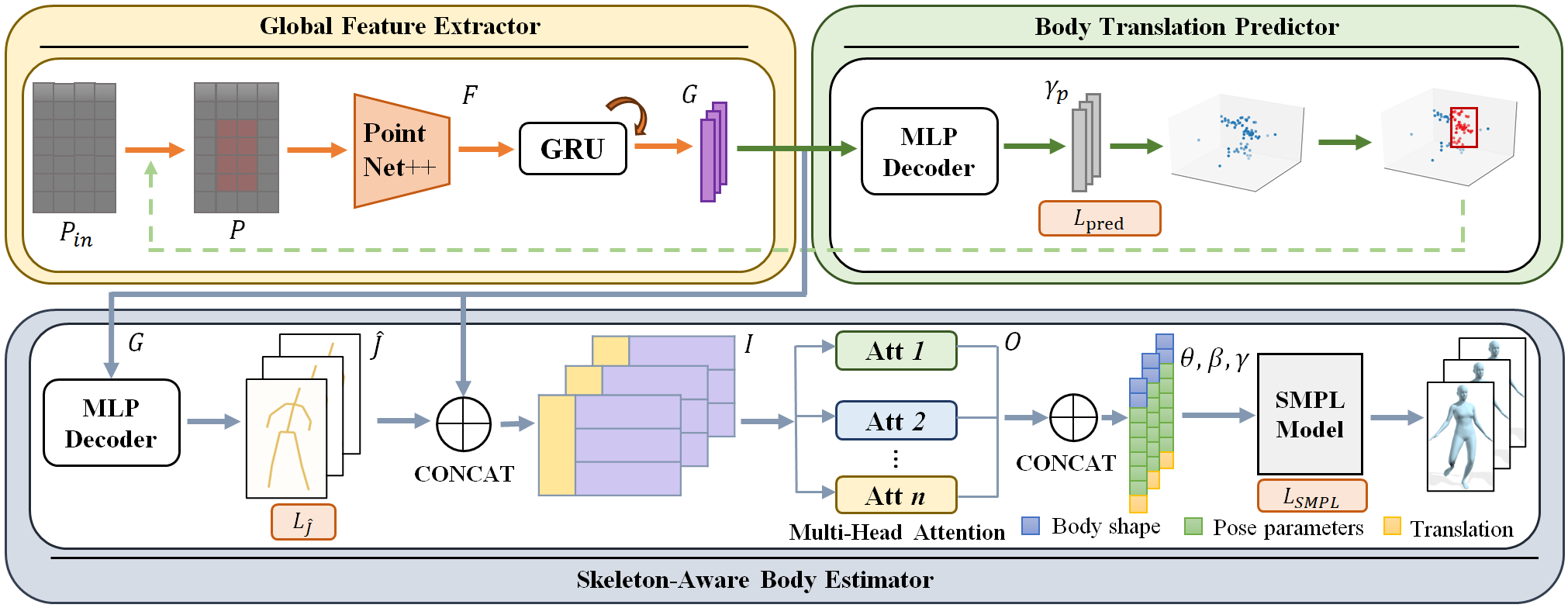}}
\caption{Overview of the proposed mmBaT framework, which comprises three modules: global feature extractor, body translation predictor, and skeleton-aware body estimator.}
\label{fig:res}
%
\end{figure*}
The contributions of the proposed method can be delineated as follows:
\par
\begin{itemize}
    \setlength{\topsep}{0pt}
    \setlength{\itemsep}{0pt}
    \setlength{\parsep}{0pt}
    \setlength{\parskip}{0pt}
    \item We propose a novel multi-task framework for human body reconstruction that takes noisy mmWave radar point clouds as input, simultaneously estimating SMPL body parameters and predicting subject translations in subsequent frames.
    \item Based on the coarse-to-fine techniques, a skeleton-aware body estimator is adopted to estimate the coarse skeleton first, which serves as prior information to effectively regress the intricate and nonlinear body parameters.
    \item With comprehensive experiments on two public datasets, our proposed method significantly outperforms the existing state-of-the-art methods, demonstrating the advancements of our method in human body reconstruction from noisy mmWave point clouds.
\end{itemize}

\section{Proposed Method}
\label{sec:format}

Fig.\ref{fig:res} provides an overview of the proposed mmBaT method, which is designed to take raw mmWave point cloud sequences \(\mathbf{P}_{\text{in}}\) as input, regardless of the number of points in each frame, and subsequently regress the SMPL pose parameter $\boldsymbol{\theta}$, shape parameter $\boldsymbol{\beta}$ and translation parameter $\boldsymbol{\gamma}$. Then they are fed into an off-the-shelf SMPL model to obtain the reconstructed 3D human joints $\mathbf{J}$ and mesh vertices $\mathbf{M}$, i.e. $\mathbf{J},\mathbf{M}=\text{SMPL}(\boldsymbol{\theta},\boldsymbol{\beta},\boldsymbol{\gamma})$.
The network architecture mainly comprises three modules: 
\textbf{Global feature extractor} aims to extract the general features that are utilized by the other two modules.
\textbf{Body translation predictor} takes the global feature as input and predicts the body translations, which are used to obtain input point clouds for the subsequent time period.
\textbf{Skeleton-aware body estimator} first estimates a coarse skeleton, and then regresses the final SMPL parameters by embedding the positions of skeleton joints with the global feature.


\subsection{Global Feature Extractor}
\label{sec:2.1}
To aggregate the spatial and temporal information, this module takes the processed point cloud sequence ${\mathbf{P}} \in \mathbb{R}^{\textit{T} \times\textit{N} \times \textit{C}}$ as input, gotten from the body translation predictor module, where \textit{N} denotes the fixed number of points in every frame and \textit{C} represents the number of channels for each point.  
\par
We make use of PointNet++ \cite{qi2017pointnet++} as the backbone to extract spatial features $\mathbf{F}$ of every point cloud frame ${\mathbf{P}}$: 
\begin{equation}
    \mathbf{F} = \text{PointNet++}(\mathbf{P})
\end{equation}
The original PointNet++ adopts a max-pooling operation to aggregate local features from the sample points obtained by furthest point sampling, which may lose information from other sampled points\cite{22}. As a result, we replace it with a linear layer inspired by the attention mechanism. Due to the inconsistency of the point clouds where some points may vanish in specific frames, the spatial features are fed into a bi-directional Gate Recurrent Unit (GRU) to obtain a 1024-dimensional global feature ${\mathbf{G}}$:
\begin{equation}
    \mathbf{G} = \text{GRU}(\mathbf{F})
\end{equation}
\par

\subsection{Body Translation Predictor}
Raw radar point clouds often contain a significant amount of noise points, which can greatly affect the accuracy of body reconstruction. 
To extract the body-related points without any specific preprocessing or prior knowledge, an intuitive approach is to select points close to the subject, but the exact locations remain elusive during the given time interval. Besides spatial and temporal information, we also use the radial velocity information inherent in the mmWave point clouds. This information serves as an indicator for the motion trend of the subject, enabling us to make predictions regarding short-term translations. 
Consequently, there is an iterative interaction between the translation predictor and the global feature extractor.
\par
Given a raw point cloud sequence \(\mathbf{P}_{\text{in}}^{[t,t+T]}\) of length $T$, 
where each frame contains an arbitrary number of points. Our objective is to select $N$ points based on the body translation \(\boldsymbol{\gamma}_{\boldsymbol{p}}^{[t,t+T]}\) predicted in the previous time window. Within this module, we take the global feature \(\mathbf{G}^{[t-T,t]}\) 
as input, and a straightforward yet effective three-layer MLP decoder is adopted here to predict \(\boldsymbol{\gamma}_{\boldsymbol{p}}^{[t,t+T]}\) for the current time window. 
The corresponding loss for this prediction is denoted as: 
\begin{equation}
\mathcal{L}_{\text{pred}}=\sum\left\|\boldsymbol{\gamma}_{GT}^{[t,t+T]}-\boldsymbol{\gamma}_{\boldsymbol{p}}^{[t,t+T]}\right\|_1
\end{equation}
\par
Given $\boldsymbol{\gamma_{p}}$ as the central point, we generate a 3D bounding box with dimensions of 1m, 1m, and 3m along the x, y, and z axes, respectively. The points contained within the bounding box will be selected, and sampling or repeating is performed as needed to ensure a uniform amount of points across frames. The generated output ${\mathbf{P}} \in \mathbb{R}^{\textit{T} \times\textit{N} \times \textit{C}}$ is the point cloud sequence, as previously mentioned in Section \ref{sec:2.1}, which is further utilized for global feature extractor. 

\subsection{Skeleton-Aware Body Estimator}
Since the SMPL model parameters are the implicit representations of the human body's pose and shape, and these parameters are highly non-linear and complex \cite{22}, direct regression from sparse and inconsistent point clouds is difficult. To address this challenge, we design the skeleton-aware body estimator that comprises two steps. 
In the first step, the coarse skeleton $\hat{\mathbf{J}} \in \mathbb{R}^{T \times N_{J} \times 3}$ is estimated by using an MLP decoder, where $N_J$ is the number of skeleton joints. 
The loss function for this step is defined as the L1 norm between the estimated coarse joint locations and the corresponding ground-truth locations:
\begin{equation}
\mathcal{L}_\mathcal{{\hat{\mathbf{J}}}}=\sum\left\|\mathbf{J}_{G T}-\hat{\mathbf{J}}\right\|_1
\label{eq:1}
\end{equation}
\par
In the second step, we aim to regress the body parameters considering the awareness of skeleton joint locations. To fuse the global information with the skeleton spatial features, we concatenate the global feature ${\mathbf{G}}$ with the joint locations $\hat{\mathbf{J}}$ and get the fused joint features $\mathbf{I} \in \ \mathbb{R}^{T\times N_{J}\times(3+1024)}$. For a better feature representation and computational efficiency, we further put $\mathbf{I}$ through a linear layer to obtain enhanced joint features $\mathbf{I^{\prime}} \in \ \mathbb{R}^{T\times N_{J}\times1024}$. Then, we employ self-attention mechanism \cite{vaswani2017attention} on $N_{J}$ to learn the implicit relationships among different joints:
\begin{align}
\text{attention}(\mathbf{Q}, \mathbf{K}, \mathbf{V}) & = \mathbf{V} \cdot \operatorname{softmax}\left(\frac{\mathbf{Q}^T \cdot \mathbf{K}}{\sqrt{\mathbf{C^k}}}\right)  \label{eq:3}
\end{align}
where $\mathbf{Q}=\mathbf{W}_Q \cdot \mathbf{I^{\prime}}$, $ \mathbf{K}=\mathbf{W}_K \cdot \mathbf{I^{\prime}}$ and $ \mathbf{V}=\mathbf{W}_V \cdot \mathbf{I^{\prime}}$ denote the obtained queries, keys and values, respectively. Then, the softmax function is performed, and the attention output is calculated as the weighted sum of the value $\mathbf{V}$.
\par
To increase the subspaces and model expressiveness, we apply multi-head attention instead of single self-attention mechanism:
\begin{align}
    \mathbf{h_{i}} & =\operatorname{attention}\left(\mathbf{Q_i} , \mathbf{K_i}, \mathbf{V_i} \right) \label{eq:4} \\
    \mathbf{O} & =\operatorname{ \mathbf{W_M} \cdot concat} \left(\mathbf{h_{1}},...,\mathbf{h_{n}}\right) \label{eq:5}
\end{align}

Multiple self-attention modules are performed in parallel with independent weights for $\mathbf{Q}$, $\mathbf{K}$ and $\mathbf{V}$. The results of $\boldsymbol{n}$ self-attention modules are concatenated to obtain the output $\mathbf{O}\in \ \mathbb{R}^{T \times N_{J}\times 1024}$ by a projection technique.
\par
At the end of this module, the output $\mathbf{O}$ is fed into an MLP layer to regress the SMPL body parameters, including the pose parameter $\boldsymbol{\theta} \in \ \mathbb{R}^{T \times 6N_{J}}$ in a 6-dimensional representation, the shape parameter $\boldsymbol{\beta}\in \ \mathbb{R}^{T \times N_{\beta}}$ and the body translation $\boldsymbol{\gamma}\in \ \mathbb{R}^{T \times 3}$. These parameters are next fed into the off-the-shelf SMPL model to compute the final joint positions $\mathbf{J}$ and mesh vertices $\mathbf{M}$. The losses associated with these parameters are defined as:
\begin{equation}
\mathcal{L}_{\text{SMPL}}=\mathcal{L}_\mathcal{\theta} + \mathcal{L}_\mathcal{\beta} + \mathcal{L}_\mathcal{\gamma}+ \mathcal{L}_\mathcal{J}+ \mathcal{L}_\mathcal{M}
\end{equation}
\par
where $\mathcal{L}_\mathcal{\beta}$, $\mathcal{L}_\mathcal{\gamma}$, $\mathcal{L}_\mathcal{J}$ and $\mathcal{L}_\mathcal{M}$ are the L1 loss with the same definition in Eq.\ref{eq:1}, while $\mathcal{L}_\mathcal{\theta}$ is the geodesic loss \cite{23} to represent the distance between the predicted and ground-truth rotation matrices. We aim to jointly optimize both tasks so the total loss of the framework is defined as:
\begin{equation}
    \mathcal{L}_{\text{total}}=\mathcal{L}_\mathcal{{\hat{\mathbf{J}}}}  + \mathcal{L}_{\text{SMPL}}+\mathcal{L}_{\text{pred}}
\end{equation}

\section{Experiments}
In this section, we provide an evaluation of our proposed method and compare its performance with prevailing state-of-the-art techniques.
\label{sec:pagestyle}
\begin{table*}[htb]
\begin{threeparttable}[b]
\centering
\caption{Comparison of Experimental Results with Other Methods}
\begin{tabular}{ccccccccccc}

\hline
\multicolumn{1}{l}{} & \multicolumn{5}{c}{mmBody \cite{chen2022mmbody}}                                                       & \multicolumn{5}{c}{MRI \cite{an2022mri}}                                                          \\ \cline{2-11} 
\multicolumn{1}{l}{} & \multicolumn{2}{c}{P4Trans \cite{fan2021p4trans}} & \multicolumn{2}{c}{mmMesh \cite{xue2021mmesh}} & \multirow{2}{*}{mmBaT} & \multicolumn{2}{c}{P4Trans \cite{fan2021p4trans}} & \multicolumn{2}{c}{mmMesh \cite{xue2021mmesh}} & \multirow{2}{*}{mmBaT} \\ \cline{2-5} \cline{7-10}
Evaluation Metrics   & w/o bbx  & w/ bbx  &   w/o bbx  & w/ bbx &      & w/o bbx  & w/ bbx  & w/o bbx   & w/ bbx  &      \\ \hline
MPJRE ($^\circ$) &      23.26     &   17.33 &  30.02 &   22.08 &    \textbf{12.83}   &  10.82  &   9.47    & 11.07&  9.94    & \textbf{9.03}             \\
MPJPE (cm)&             39.65  &  33.25   & 32.26    & 28.91  &  \textbf{25.10}   &           12.98  &  10.39  & 12.46  &  11.55    &   \textbf{9.73}         \\
MPVPE (cm)&             31.04   & 25.95   & 25.17 &  21.60 &    \textbf{19.91}   &  /        &      /    &      /   &   / & /  \\
MTE (cm) &   18.13 &  16.81  & 20.43 & \textbf{15.46} &   16.13 &  8.46   &  8.04  &  8.41  &    8.25   & \textbf{7.76}   \\
MPTE (cm) &              -    &   -     &    -         &    -   &      \textbf{14.54}           &        -      &          -    &        -     &         -     &     \textbf{8.16}               \\ 

\hline
\label{tab:1}
\end{tabular}
\vspace{-0.5em}
 \begin{tablenotes}
 \footnotesize 
 \item Note: MPVPE is not compared in MRI dataset since mesh vertices are not provided. MPTE is specifically designed for our method, and it is not valid for others.
 \end{tablenotes}
\end{threeparttable}
\vspace{-0.5em}
\end{table*}
\vspace{-0.5em}

\subsection{Dataset Descriptions}
We employ two publicly available datasets to evaluate our proposed method:

\textbf{mmBody dataset} \cite{chen2022mmbody} is the first public dataset that offers Motion Capture (MoCap) ground-truth data in the SMPL-X format \cite{pavlakos2019smplx} for mmWave-based human body reconstruction task. It contains more than 100 motions performed by 20 subjects and covers different environments like dark, rain, and fog. The radar device has a range resolution of 0.4 cm, with each frame comprising thousands of points, the majority of which are noise points. 
\par
\textbf{MRI dataset} \cite{an2022mri} mainly focuses on rehabilitation exercises, which contains over 160k frames from 20 subjects in a house monitoring condition. The radar device used in this dataset has a lower range resolution compared to \cite{chen2022mmbody}, each frame has only 64 points. 
A method was presented in this paper to augment the number of points by fusing data from three consecutive frames. Since ground-truth data only contains 17 joint locations, mesh vertices are not considered for this dataset.
\subsection{Experimental Setup}
In this part, we will introduce our experimental setup in detail from the following aspects:

\textit{1) Training details:} The deep learning architecture is implemented with PyTorch and trained on an NVIDIA GeForce RTX 3090 GPU. We train our method for 100 epochs with an Adam optimizer. The dropout ratio is set to 0.2 for GRU layers and linear layers. The initial learning rate is set to $1 \times 10^{-3}$ with a decay rate $1 \times 10^{-4}$. The batch size is set to 16. We apply the scale factors to normalize each loss term to the same scale. The number of points sampled from each frame is 1024 and 196 for mmBody dataset and MRI dataset, respectively. The training set and test set are divided according to the original dataset, and \( 10 \% \) of the training data is used for validation. An initial bounding box is provided for our method for every test sequence. 

\textit{2) Compared methods:} A thorough evaluation of our proposed method is conducted on two existing methods: \textbf{P4Transformer} (\textbf{P4Trans}) \cite{fan2021p4trans} and \textbf{mmMesh} \cite{xue2021mmesh}. Specifically, P4Transformer is the benchmark employed to assess the validity of the mmBody dataset \cite{chen2022mmbody}. To further evaluate the performance of our body estimator module, besides the original implementations of these two methods, without given ground-truth bounding boxes (w/o bbx), we also compare our method with the modified version of these two methods provided with ground-truth bounding boxes (w/ bbx). 
This configuration is adopted to exclude the influence of noisy point clouds.

\textit{3) Evaluation metrics:} 
\par
\textbf{Mean Per Joint Rotation Error (MPJRE)} \cite{xue2021mmesh} calculates the average differences between predicted joint rotations and the ground-truth rotations, which reports the pose accuracy of reconstruction results. 

\textbf{Mean Per Joint Position Error (MPJPE)} \cite{5, 7} calculates Euclidean distance between the estimated and ground-truth joint positions.

\textbf{Mean Per Vertice Position Error (MPVPE)} \cite{7, bogo2016SMPL} calculates Euclidean distance between the estimated and ground-truth error positions. This metric reports the overall mesh reconstruction results, which are affected by pose error, location error, and shape error.

\textbf{Mean Translation Error (MTE)} \cite{xue2021mmesh} measures Euclidean distance between predicted and ground-truth locations of the skeleton root joint. This metric represents the precision of body localization. 

\textbf{Mean Prediction Translation Error (MPTE)} measures Euclidean distance between predicted and ground-truth locations of the skeleton root joint in our body translation predictor module. 


\subsection{Experimental Results and Analysis}

\begin{figure}[htb]
\centering
\centerline{\includegraphics[width=8.5cm]{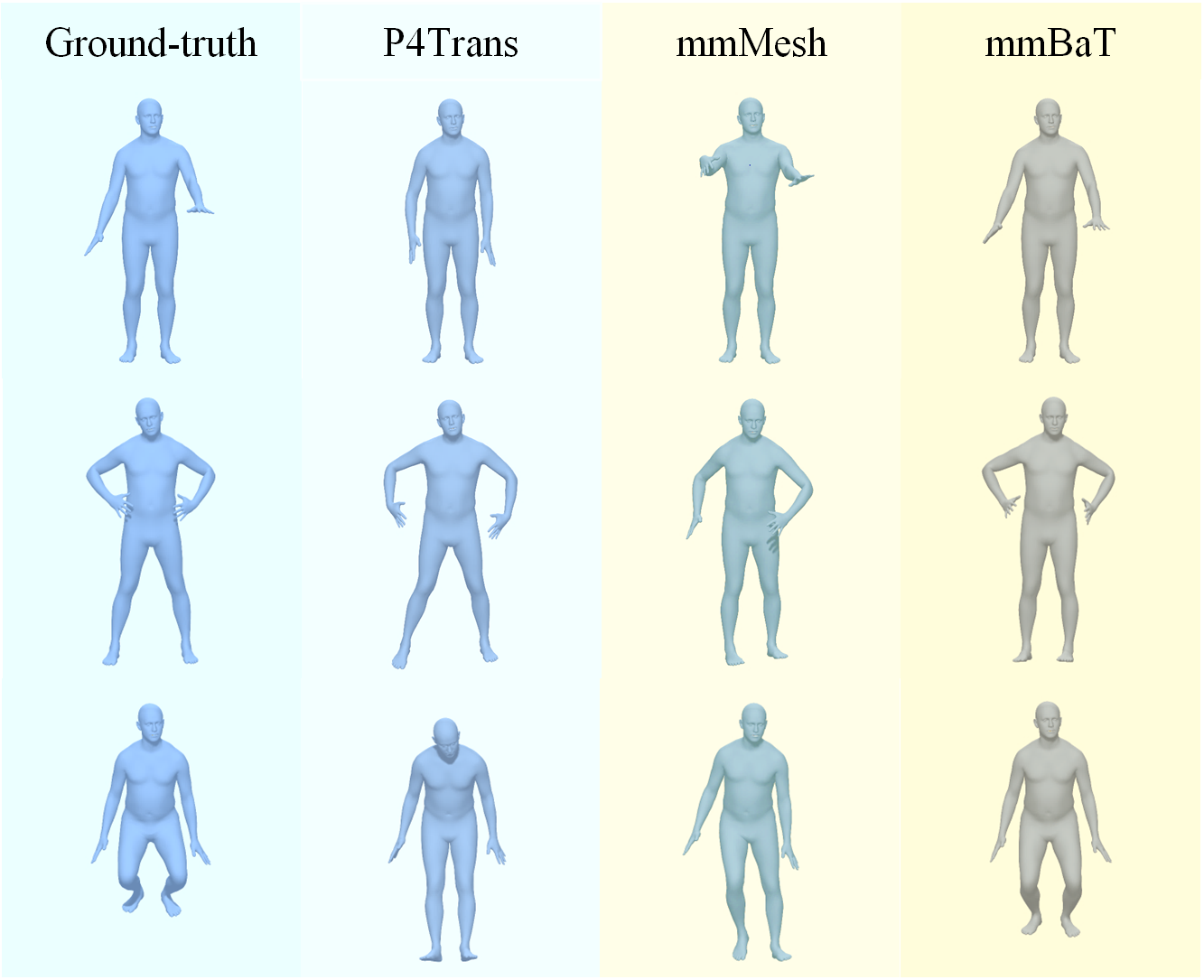}}
\caption{Qualitative comparisons between mmBaT and baselines from mmBody dataset.}
\label{fig:2}
\end{figure}
In this section, we present the results of our comprehensive experiments, which evaluate the performance of our proposed mmBaT against existing methods. The quantitative results of our experiments are displayed in Table \ref{tab:1}. 

\textit{1) Quantitative analysis:} Overall, mmBaT consistently outperforms other methods across a range of metrics, with notable strengths observed in MPJRE, MPJPE, and MPVPE. These metrics are indicative of the superior ability of our method to accurately reconstruct human body poses and shapes. 
In particular, the significant improvement in the pose error (MPJRE) shows an average enhancement of 4.5$^\circ$ and 9.25$^\circ$ compared to P4Transformer and mmMesh in mmBody dataset, with ground-truth bounding box implementation. This significant improvement can be attributed to our coarse-to-fine approach, in which pose parameters are regressed based on the coarse skeleton as a priori, revealing the efficacy of our body estimator module in addressing those challenges. In terms of MTE, our approach closely matches the optimal results, deviating by only 0.7cm on average. This proximity to the optimal results is largely facilitated by the ground-truth bounding box, which aids in accurate subject localization.

By comparing the two methods with and without bounding box settings, we see great improvements across all metrics on the mmBody dataset, demonstrating the critical role of body translation prediction in accurate body reconstruction.  
Moreover, our method consistently achieves an MPTE under 15cm, showcasing its effectiveness in accurately predicting body translations and adeptly mitigating the influence of noisy point clouds.
\par

In MRI dataset, all methods have relatively good performance, mainly because MRI dataset involves simple rehabilitation motions captured on a small scale and the point clouds contain fewer noise points. While mmBaT still remains the best results across all metrics, which shows the robustness of our approach in different environments.


\textit{2) Qualitative analysis:} In Fig.2, we provide a visual demonstration of body reconstruction results, using mmBody dataset as a reference. The figure presents ground-truth data alongside the reconstruction results produced by two comparative methods with the ground-truth bounding box, followed by the outcomes generated by our proposed method, progressing sequentially from left to right. To facilitate a comprehensive evaluation of human poses, three representative frames have been selected. It can be visualized that our method achieves more favorable reconstruction results compared to other algorithms for pose estimation, proving the effectiveness of our novel multi-task framework and the superiority of our method.

\par 
\par 


\section{Conclusion}
In this paper, we introduced mmBaT, a novel multi-task framework that simultaneously enhances the accuracy of human body reconstruction and facilitates the extraction of body-related points from noisy mmWave point clouds. 
It was developed with the specific goal of addressing the persistent challenges posed by noisy mmWave point clouds and the inherent difficulties in achieving accurate pose estimations, which have remained unresolved by existing methods.
By adopting a coarse-to-fine manner, our framework showed a remarkable performance improvement in the accuracy of human pose estimation, with gains up to 9.25$^\circ$ observed. Our approach has been evaluated on two public datasets, demonstrating its robustness and effectiveness comprehensively. Experimental results proved the superiority of our method in providing accurate and reliable outcomes for human body reconstruction in challenging environments.
\clearpage
\vfill\pagebreak

\bibliographystyle{IEEEbib}
\balance
\bibliography{strings,refs}

\begin{thebibliography}{10}

\bibitem{1}
Marco Mercuri, Ilde~Rosa Lorato, Yao-Hong Liu, Fokko Wieringa, Chris~Van Hoof,
  and Tom Torfs,
\newblock ``Vital-sign monitoring and spatial tracking of multiple people using
  a contactless radar-based sensor,''
\newblock {\em Nature Electronics}, vol. 2, no. 6, pp. 252--262, 2019.

\bibitem{2}
Kawon Han and Songcheol Hong,
\newblock ``Cough detection using millimeter-wave fmcw radar,''
\newblock in {\em ICASSP 2023-2023 IEEE International Conference on Acoustics,
  Speech and Signal Processing (ICASSP)}. IEEE, 2023, pp. 1--5.

\bibitem{xia2022boundary}
Songpengcheng Xia, Lei Chu, Ling Pei, Wenxian Yu, and Robert~C Qiu,
\newblock ``A boundary consistency-aware multitask learning framework for joint
  activity segmentation and recognition with wearable sensors,''
\newblock {\em IEEE Transactions on Industrial Informatics}, vol. 19, no. 3,
  pp. 2984--2996, 2022.

\bibitem{3}
Hoang~Thanh Le, Son~Lam Phung, Abdesselam Bouzerdoum, and Fok Hing~Chi Tivive,
\newblock ``Human motion classification with micro-doppler radar and
  bayesian-optimized convolutional neural networks,''
\newblock in {\em 2018 IEEE International Conference on Acoustics, Speech and
  Signal Processing (ICASSP)}. IEEE, 2018, pp. 2961--2965.

\bibitem{5}
Wenjun Jiang, Hongfei Xue, Chenglin Miao, Shiyang Wang, Sen Lin, Chong Tian,
  Srinivasan Murali, Haochen Hu, Zhi Sun, and Lu~Su,
\newblock ``Towards 3d human pose construction using wifi,''
\newblock in {\em Proceedings of the 26th Annual International Conference on
  Mobile Computing and Networking}, 2020, pp. 1--14.

\bibitem{7}
Mingmin Zhao, Tianhong Li, Mohammad Abu~Alsheikh, Yonglong Tian, Hang Zhao,
  Antonio Torralba, and Dina Katabi,
\newblock ``Through-wall human pose estimation using radio signals,''
\newblock in {\em Proceedings of the IEEE conference on computer vision and
  pattern recognition}, 2018, pp. 7356--7365.

\bibitem{tian2023recovering}
Yating Tian, Hongwen Zhang, Yebin Liu, and Limin Wang,
\newblock ``Recovering 3d human mesh from monocular images: A survey,''
\newblock {\em IEEE Transactions on Pattern Analysis and Machine Intelligence},
  2023.

\bibitem{pei2021mars}
Ling Pei, Songpengcheng Xia, Lei Chu, Fanyi Xiao, Qi~Wu, Wenxian Yu, and Robert
  Qiu,
\newblock ``Mars: Mixed virtual and real wearable sensors for human activity
  recognition with multidomain deep learning model,''
\newblock {\em IEEE Internet of Things Journal}, vol. 8, no. 11, pp.
  9383--9396, 2021.

\bibitem{sengupta2020mm}
Arindam Sengupta, Feng Jin, Renyuan Zhang, and Siyang Cao,
\newblock ``mm-pose: Real-time human skeletal posture estimation using mmwave
  radars and cnns,''
\newblock {\em IEEE Sensors Journal}, vol. 20, no. 17, pp. 10032--10044, 2020.

\bibitem{xue2021mmesh}
Hongfei Xue, Yan Ju, Chenglin Miao, Yijiang Wang, Shiyang Wang, Aidong Zhang,
  and Lu~Su,
\newblock ``mmmesh: Towards 3d real-time dynamic human mesh construction using
  millimeter-wave,''
\newblock in {\em Proceedings of the 19th Annual International Conference on
  Mobile Systems, Applications, and Services}, 2021, pp. 269--282.

\bibitem{bogo2016SMPL}
Federica Bogo, Angjoo Kanazawa, Christoph Lassner, Peter Gehler, Javier Romero,
  and Michael~J Black,
\newblock ``Keep it smpl: Automatic estimation of 3d human pose and shape from
  a single image,''
\newblock in {\em Computer Vision--ECCV 2016: 14th European Conference,
  Amsterdam, The Netherlands, October 11-14, 2016, Proceedings, Part V 14}.
  Springer, 2016, pp. 561--578.

\bibitem{xue2022m4esh}
Hongfei Xue, Qiming Cao, Yan Ju, Haochen Hu, Haoyu Wang, Aidong Zhang, and
  Lu~Su,
\newblock ``M4esh: mmwave-based 3d human mesh construction for multiple
  subjects,''
\newblock in {\em Proceedings of the 20th ACM Conference on Embedded Networked
  Sensor Systems}, 2022, pp. 391--406.

\bibitem{chen2023immfusion}
Anjun Chen, Xiangyu Wang, Kun Shi, Shaohao Zhu, Bin Fang, Yingfeng Chen, Jiming
  Chen, Yuchi Huo, and Qi~Ye,
\newblock ``Immfusion: Robust mmwave-rgb fusion for 3d human body
  reconstruction in all weather conditions,''
\newblock in {\em 2023 IEEE International Conference on Robotics and Automation
  (ICRA)}. IEEE, 2023, pp. 2752--2758.

\bibitem{22}
Haiyong Jiang, Jianfei Cai, and Jianmin Zheng,
\newblock ``Skeleton-aware 3d human shape reconstruction from point clouds,''
\newblock in {\em Proceedings of the IEEE/CVF International Conference on
  Computer Vision}, 2019, pp. 5431--5441.

\bibitem{cho2021rethinking}
Sung-Jin Cho, Seo-Won Ji, Jun-Pyo Hong, Seung-Won Jung, and Sung-Jea Ko,
\newblock ``Rethinking coarse-to-fine approach in single image deblurring,''
\newblock in {\em Proceedings of the IEEE/CVF international conference on
  computer vision}, 2021, pp. 4641--4650.

\bibitem{sarlin2019coarse}
Paul-Edouard Sarlin, Cesar Cadena, Roland Siegwart, and Marcin Dymczyk,
\newblock ``From coarse to fine: Robust hierarchical localization at large
  scale,''
\newblock in {\em Proceedings of the IEEE/CVF Conference on Computer Vision and
  Pattern Recognition}, 2019, pp. 12716--12725.

\bibitem{qi2017pointnet++}
Charles~Ruizhongtai Qi, Li~Yi, Hao Su, and Leonidas~J Guibas,
\newblock ``Pointnet++: Deep hierarchical feature learning on point sets in a
  metric space,''
\newblock {\em Advances in neural information processing systems}, vol. 30,
  2017.

\bibitem{vaswani2017attention}
Ashish Vaswani, Noam Shazeer, Niki Parmar, Jakob Uszkoreit, Llion Jones,
  Aidan~N Gomez, {\L}ukasz Kaiser, and Illia Polosukhin,
\newblock ``Attention is all you need,''
\newblock {\em Advances in neural information processing systems}, vol. 30,
  2017.

\bibitem{23}
Siddharth Mahendran, Haider Ali, and Ren{\'e} Vidal,
\newblock ``3d pose regression using convolutional neural networks,''
\newblock in {\em Proceedings of the IEEE International Conference on Computer
  Vision Workshops}, 2017, pp. 2174--2182.

\bibitem{chen2022mmbody}
Anjun Chen, Xiangyu Wang, Shaohao Zhu, Yanxu Li, Jiming Chen, and Qi~Ye,
\newblock ``mmbody benchmark: 3d body reconstruction dataset and analysis for
  millimeter wave radar,''
\newblock in {\em Proceedings of the 30th ACM International Conference on
  Multimedia}, 2022, pp. 3501--3510.

\bibitem{an2022mri}
Sizhe An, Yin Li, and Umit Ogras,
\newblock ``mri: Multi-modal 3d human pose estimation dataset using mmwave,
  rgb-d, and inertial sensors,''
\newblock {\em Advances in Neural Information Processing Systems}, vol. 35, pp.
  27414--27426, 2022.

\bibitem{fan2021p4trans}
Hehe Fan, Yi~Yang, and Mohan Kankanhalli,
\newblock ``Point 4d transformer networks for spatio-temporal modeling in point
  cloud videos,''
\newblock in {\em Proceedings of the IEEE/CVF conference on computer vision and
  pattern recognition}, 2021, pp. 14204--14213.

\bibitem{pavlakos2019smplx}
Georgios Pavlakos, Vasileios Choutas, Nima Ghorbani, Timo Bolkart, Ahmed~AA
  Osman, Dimitrios Tzionas, and Michael~J Black,
\newblock ``Expressive body capture: 3d hands, face, and body from a single
  image,''
\newblock in {\em Proceedings of the IEEE/CVF conference on computer vision and
  pattern recognition}, 2019, pp. 10975--10985.

\end{thebibliography}

\end{document}